\documentclass[5p,twocolumn,authoryear]{elsarticle}

\usepackage{amssymb}
\usepackage{amsmath}

\usepackage{lineno}
\usepackage{caption}
\usepackage{multirow}
\usepackage{booktabs}
\usepackage{url}
\usepackage{verbatim}
\usepackage{orcidlink}
\usepackage{wasysym}
\usepackage[caption=false,font=normalsize,labelfont=sf,textfont=sf]{subfig}

\journal{Nuclear Physics B}

\begin{document}

\begin{frontmatter}



\title{Multi-Hazard Early Warning Systems for Agriculture \\with Featural-Temporal Explanations\\\small{(Pre-print v0.8 2025-07-30)}} 


\author[UTS]{Boyuan Zheng\,\orcidlink{0000-0003-1223-9230}} 
\author[UTS]{Victor W. Chu\,\orcidlink{0000-0002-5853-5820}} 

\affiliation[UTS]{organization={A-Theme, Data Science Institute, University of Technology Sydney},
            addressline={15 Broadway, Ultimo}, 
            city={Sydney},
            postcode={2007}, 
            state={NSW},
            country={Australia}}

\begin{abstract}
Climate extremes present escalating risks to agriculture intensifying the need for reliable multi-hazard early warning systems (EWS). The situation is evolving due to climate change and hence such systems should have the intelligent to continue to learn from recent climate behaviours.  However, traditional single-hazard forecasting methods fall short in capturing complex interactions among concurrent climatic events. To address this deficiency, in this paper, we combine sequential deep learning models and advanced Explainable Artificial Intelligence (XAI) techniques to introduce a multi-hazard forecasting framework for agriculture. In our experiments, we utilize meteorological data from four prominent agricultural regions in the United States (between 2010 and 2023) to validate the predictive accuracy of our framework on multiple severe event types, which are extreme cold, floods, frost, hail, heatwaves, and heavy rainfall, with tailored models for each area. The framework uniquely integrates attention mechanisms with TimeSHAP (a recurrent XAI explainer for time series) to provide comprehensive temporal explanations  revealing not only which climatic features are influential but precisely when their impacts occur. Our results demonstrate strong predictive accuracy, particularly with the BiLSTM architecture, and highlight the system’s capacity to inform nuanced, proactive risk management strategies. This research significantly advances the explainability and applicability of multi-hazard EWS, fostering interdisciplinary trust and effective decision-making process for climate risk management in the agricultural industry.
\end{abstract}




\end{frontmatter}


\section{Introduction}
\label{sec:intro}

Agriculture is known to be vulnerable to climate change, which has already impacted crop production in a measurable manner \citep{schlenker2009nonlinear, lobell2025half,betts2018changes}. In particular, extreme weather events are amplifying such problem and pose escalating threats to agricultural productivity and economic stability, especially as climate change drives an increase in the frequency and intensity of climate extremes worldwide \citep{lesk2016influence,chen2018projected}. Hazards, such as extreme cold, floods, frost, hail, heatwaves, and heavy rainfall, can devastate crop yields and supply chains incurring massive financial losses \citep{liu2020impacts, schmitt2022extreme}. In recent decades, climate-related disasters have cost hundreds of billions in damages, and recent years have seen a disproportionate share of these losses as extreme events become 
widespread \citep{powell2016measuring}. Early warning systems (EWS) are therefore 
becoming necessities to the agricultural sector and broader economy, providing advance hazard forecasts that enable farmers, communities, and decision-makers to take proactive measures to mitigate adverse climate impacts \citep{basher2006global, reichstein2025early,camps2025artificial}.

However, surging climate and weather extremes increasingly involve multiple, concurrent, or cascading hazards, which pose 
unprecedented challenges to traditional single-focus EWS \citep{brett2025science}. An intense weather episode may trigger various threats in tandem (e.g., heavy rain leading to floods and crop disease, or a late spring frost following an early warm spell), stressing the need for integrated monitoring of diverse perils. Recognizing this complexity, the international community has emphasized the development of multi-hazard early warning systems 
as a priority.  For example, the United Nations and World Meteorological Organization launched the “Early Warnings for All” initiative\footnote{\url{https://www.un.org/en/climatechange/early-warnings-for-all}} in 2022 to expand EWS coverage for all major hazards \citep{WMO2022}. A unified multi-hazard EWS offers the versatility of consolidating forecasts for different extremes within a single framework, which is especially valuable for agricultural risk management. By simultaneously predicting multiple hazard types (rather than operating separate siloed warning tools for each peril), a unified approach can provide more comprehensive situational awareness and efficient resource allocation for disaster preparedness \citep{zhang2023machine}. In practice, state-of-the-art forecasting methods have mostly remained hazard-specific — for instance, specialized data-driven models exist for individual perils such as floods, droughts, or hailstorms \citep{beillouin2020impact, lobell2011climate} — yet comparatively few attempts have been made to develop integrated multi-hazard systems \citep{hrast2025expert, reichstein2025early}. This 
negligence highlights the need for an early warning solution capable of handling a suite of hazards in tandem, thereby improving the coordination of protective actions across different threat scenarios.

On the other hand, the agricultural domain is inherently interdisciplinary: meteorologists, agronomists, and policy-makers must collaborate and develop trust on common platforms. While black-box machine learning models are powerful forecasters, they often lack of transparency impeding their adoption in high-stakes decision environments \citep{zhou2021evaluating, arrieta2020explainable}. Explainable Artificial Intelligence (XAI) techniques seek to address this issue by illuminating the reasoning behind model predictions, thereby enhancing users' trust, accountability, and uptake of AI-driven tools \citep{arrieta2020explainable}. In the context of hazard forecasting, XAI can bolster confidence in an EWS by clarifying why a model predicts an extreme event.  For example, \cite{budimir2025opportunities} identified the atmospheric signals or precursor conditions that led to alerts in his recent work to provide explanation. Such transparency is crucial for interdisciplinary trust: stakeholders from different fields can verify that the model’s behavior aligns with domain knowledge and can more readily act on its warnings. Modern studies have accordingly begun integrating XAI into climate risk tools, with notable success in hazard-specific applications (e.g., AI models augmented with explainability to detect wildfires, landslides, or floods) \citep{hrast2025expert, cilli2022explainable}.
Nevertheless, most current XAI applications in time-series hazard forecasting tend to focus on feature-level explanations, emphasizing which input features (e.g., temperature, precipitation indices) most influenced a prediction \citep{hrast2025expert, cilli2022explainable}. These feature attributions, while informative, do not capture the temporal development of extreme events — that is, they tell what factors mattered but not when. For sequential models that ingest meteorological time series, understanding the timing and evolution of predictive signals is vital. For instance, decision-makers may wish to know whether an impending drought warning was driven by anomalous rainfall patterns in preceding months or by a sudden recent weather shift. However, typical feature importance analyses (e.g., SHAP value rankings of features) lack a temporal dimension and thus cannot reveal how early-season conditions versus immediate precursors contributed to the hazard forecast. This limitation in existing XAI research motivates the pursuit of richer, temporal explanations that elucidate the sequence of events leading to a forecast.

\begin{figure*}[ht]
    \centering
    \includegraphics[width=0.82\linewidth]{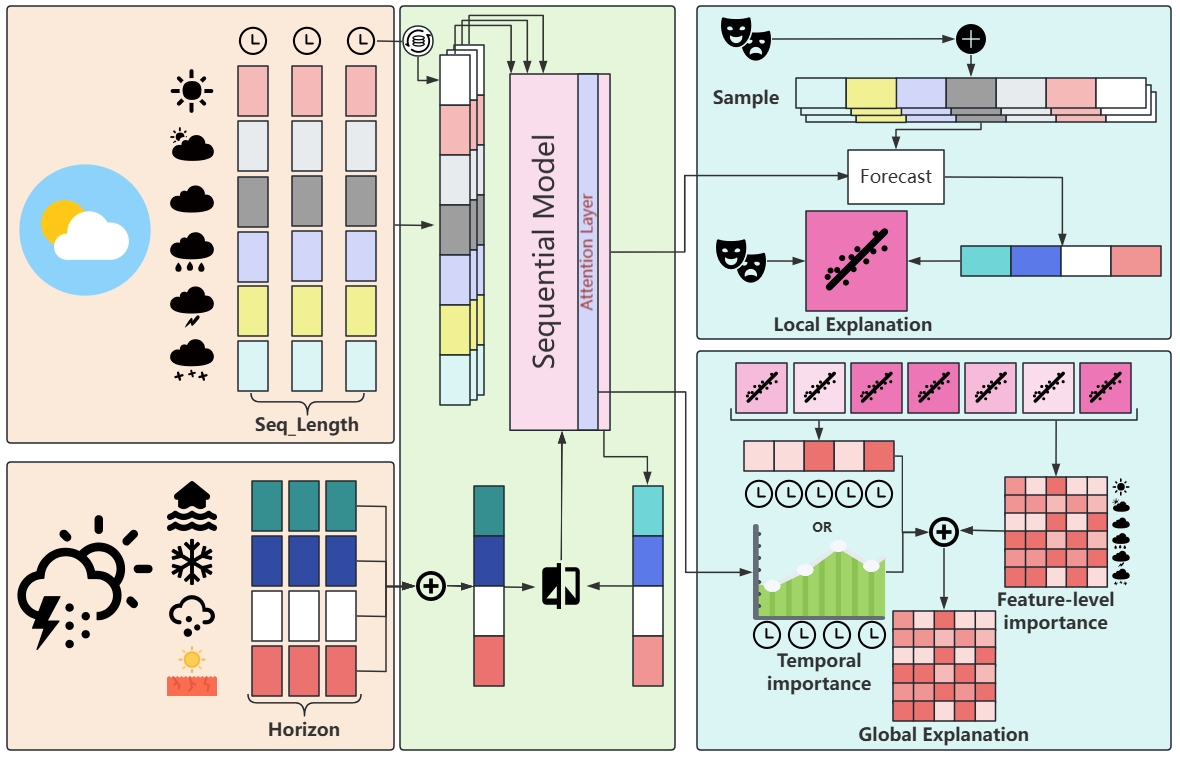}
    \caption{Graphical representation of the unified multi-hazard forecasting framework for agriculture. The diagram illustrates data integration from meteorological sources, sequential model architecture with attention mechanisms, and the dual-layered explainability approach employing TimeSHAP for both local and global hazard explanations.}
    \label{fig:framework}
\end{figure*}

To address these deficiencies, we develop a unified multi-hazard early warning system for agriculture that not only forecasts multiple extreme events but also provides explainable temporal insights into each prediction as shown in Figure~\ref{fig:framework}. Our framework leverages sequential deep learning models with an attention mechanism, to learn patterns from a decade of weather station data (2010-2023) from the United States and to provide 
probabilistic forecasts for six hazard types: extreme cold, floods, frost, hail, heat, and heavy rainfall. The sequential models allows the system to capture complex temporal dependencies in meteorological data. Moreover, we embed explainability at the core of the system. The attention mechanism provides an intrinsic form of temporal explanation, by weighting the importance of past time steps when making each prediction. In addition, we adapt TimeSHAP \citep{bento2021timeshap} — a model-agnostic XAI method extending SHAP \citep{lundberg2017unified} to sequential data — to generate post-hoc explanations that quantify the contribution of each time step (and each feature) to a given warning. TimeSHAP computes feature-level and timestep-level attributions for recurrent models, thereby identifying not only which features but also which time points or periods were most influential in the model’s decision. By combining attention-based explainability with TimeSHAP analyses, our system yields robust temporal explanations for every hazard forecast, helping to reveal the seasonal and short-term precursors that drive extreme agricultural events. We posit that this unified, explainable multi-hazard EWS will enhance transparency and interdisciplinary trust in the model outputs, ultimately improving the system’s effectiveness as a decision-support tool for climate resilience and agricultural risk management.

The structure of this paper is organized as follows:
Section~\ref{sec:intro} introduces the motivation for developing a unified, explainable multi-hazard early warning system in agriculture.
Section~\ref{sec:related_work} reviews relevant literature on early warning systems and 
XAI for climate hazard forecasting.
Section~\ref{sec:data} describes the data sources, preprocessing steps, and selection of agricultural regions used in this study.
Section~\ref{sec:methodology} outlines the sequential modeling framework, training 
mechanism, and the integration of 
explainability techniques.
Section~\ref{sec:results} presents forecasting outcomes and explainability analyses, including both local and global temporal explanations.
Finally, Section~\ref{sec:con} concludes the study with a comprehensive summary.

\section{Related Work}
\label{sec:related_work}
Research on early warning systems (EWS) has gained renewed momentum since the launch of the United Nations' “Early Warnings for All” initiative in 2022 \citep{WMO2022}. In this context, \cite{abdalzaher2023employing} proposed an architecture that integrates Internet-of-Things sensors with machine learning techniques to enhance earthquake early warning systems in urban environments. Their framework aims to process real-time seismic signals more efficiently, thereby improving alert responsiveness in smart cities. Yet, the effectiveness of EWS is not solely determined by technological advances. As shown by \cite{shah2023community}, social and institutional challenges — such as limited trust in alerts, poor dissemination practices, and the marginalization of vulnerable groups — can undermine the utility of flood warnings in regions like Pakistan. Their findings underscore the importance of NGO-led communication strategies to fill these gaps. Moving into 2025, attention has increasingly turned to multi-hazard and integrated systems. For instance, \cite{reichstein2025early} presented a framework that couples AI-driven climate forecasting with impact assessments, advocating for user-focused design and transparent model governance to address complex climate risks more proactively. Complementing this direction, \cite{rokhideh2025multi} assessed EWS progress within the Sendai Framework, identifying structural limitations such as fragmented hazard coverage, insufficient funding in the Global South, and inconsistent terminology across platforms. They call for more inclusive and coordinated systems that merge top-down policies with local engagement. In agriculture, \cite{de2025towards} advanced an operations research-based model for anticipatory action, using optimization methods to activate timely interventions — such as input distribution or early financial assistance — based on predictive agricultural hazard signals. This approach offers an alternative to rigid threshold-based warnings and has shown improved responsiveness in farming contexts.

In parallel with advances in early warning systems, the field has witnessed significant progress in explainable artificial intelligence (XAI), particularly in the context of time-series modeling. \cite{nayebi2023windowshap} proposed WindowSHAP, a method that segments time series to efficiently estimate Shapley values, thereby identifying key temporal regions driving a model’s output while keeping computational demands manageable. In a related effort, \cite{zheng2024explaining} addressed sequential decision-making by introducing R2RISE, a technique that highlights influential frames in imitation learning tasks, offering insights into how policies respond to specific temporal contexts. Within meteorology, XAI methods have begun to play a more prominent role in enhancing the explainability of hazard prediction systems. For example, \cite{cilli2022explainable} applied an explainable random forest model to identify wildfire risk in Southern Europe, leveraging feature attribution to uncover major contributing factors — such as drought indicators and vegetation health — and to delineate regions of elevated concern. Addressing a broader range of climate hazards, \cite{hrast2025expert} designed an XAI framework guided by domain expertise, where models trained on historical agro-climatic data produce probabilistic warnings for extreme events like heatwaves and droughts, accompanied by explainable justifications that align with expert reasoning. Despite these advancements, concerns have been raised about the limits of post hoc explainability. \cite{o2025moving} argue for a shift toward inherently transparent modeling approaches, drawing inspiration from the modular structure of physics-based climate simulations, where each component can be systematically evaluated. They suggest that embedding explainability directly into the model architecture offers deeper and more actionable understanding than retrospective explanations. Complementing these perspectives, \cite{pathania2025interpretable} developed a transformer-based drought forecasting model tailored for India, incorporating attention mechanisms to reveal which spatial and temporal inputs are most critical to the prediction. Their work demonstrates that high-performing deep learning models can still yield explainable outputs, thereby facilitating informed decision-making in climate-sensitive domains. Collectively, this body of work reflects a growing emphasis on not only achieving accurate multi-hazard forecasts but also ensuring their outputs are understandable and actionable for stakeholders in agriculture and environmental planning.

\section{Data}
\label{sec:data}


For this study, we select climate data from four counties of the United States: i) Sonoma (California), ii) Kent (Michigan), iii) Adams (Pennsylvania), and iv) Yakima (Washington).  They are widely recognized for their significant contributions to regional agriculture. These locations are among the leading fruit-producing areas in their respective states referring to \citet{usdadata}. Daily weather observations at the county level, spanning from 2010 to 2023, were collected from nearby NOAA weather stations using the NCEI \citet{noaaclimate} (CDO) platform. Station selection prioritized feature completeness and temporal coverage. The meteorological dataset includes features related to temperature (e.g., daily highs and lows), precipitation, snowfall, and sunlight exposure. To capture weather-related hazards, we additionally sourced historical records of extreme events from NOAA’s \citet{noaaee} covering the same period. This dataset includes event type, severity, and duration. We excluded events of light or moderate impact based on associated economic losses, agricultural damage, and casualties, retaining only six categories of severe events: Extreme Cold, Flood, Frost, Hail, Heat, and Extreme Rain. Droughts were excluded due to their infrequent reporting in the selected regions.

Prior to modeling, several preprocessing steps were applied to prepare the data. Meteorological features with substantial missing values were discarded, and event records were aggregated to the county level to ensure spatial consistency with the climate data. The six severe event types were converted into binary indicators through one-hot encoding, denoting event occurrence on a daily basis. The weather and event datasets were then temporally aligned and merged. A 14-day forecasting window was adopted: for each observation day, the number of severe events occurring within the following 14 days was summed by type to construct multi-label prediction targets. Lastly, all continuous meteorological features were standardized using a zero-mean, unit-variance transformation 
to facilitate model training.

\section{Methodology}
\label{sec:methodology}

The forecasting framework presented in this research integrates several sequential models enhanced by attention mechanisms to simultaneously predict multiple agricultural hazards (as shown in Figure~\ref{fig:framework}). Our approach employs a many-to-one structure, outputting predictions representing the anticipated occurrence of extreme events within a 90-day forecasting horizon. Daily meteorological inputs, gathered from weather stations distributed across the United States, are formatted as two-dimensional arrays, reflecting time steps and climate features, and expand into three-dimensional arrays when batch processing is used. Model performance and stability are rigorously assessed across various sequence lengths and stride sizes to optimize temporal coverage and resolution, thereby enhancing predictive precision. 

We employ the Poisson Negative Log Likelihood (NLL) as loss function in model training.  It was chosen for its suitability in modeling the nature of our target variables — discrete event counts such as the number of frost, hail, or drought occurrences over a given period. Unlike standard regression losses that assume continuous and normally distributed targets, Poisson NLL more accurately captures the sparsity and stochasticity typical of agricultural hazard data \citep{nelder1974log}. For optimization, we use the AdamW algorithm with a learning rate of $1 \times 10^{-3}$. AdamW is selected for its ability to decouple weight decay from the gradient update, offering more effective regularization \citep{parikh2014proximal}. This is particularly advantageous given the diversity of regional climates and the sequential nature of the input data, where overfitting is a common concern. Its stable convergence characteristics and generalization performance further support its application in our deep learning setup.

Several model architectures are evaluated, including Long Short-Term Memory (LSTM), Bidirectional LSTM (BiLSTM), and Transformer-based methods. Attention mechanisms, specifically Bahdanau attention layers, are integrated into the LSTM and BiLSTM models to effectively capture critical temporal dependencies within input sequences 
(please refer to \citet{zhang2023dive} for details).
Mathematically, given a sequence $X = {x_1, x_2, \dots, x_T}$, attention-weighted representations are computed through $\alpha_t = \text{softmax}(\text{score}(h_t, s_{T}))$ and aggregated as $c = \sum_{t=1}^{T}\alpha_t h_t$, where $h_t$ denotes the hidden state at time $t$, $s_{T}$ the sequence’s final state, and $\alpha_t$ the associated attention weights. The Transformer model incorporates multi-head self-attention mechanisms and position-wise feed-forward networks, deliberately configured with compact embedding dimensions and fewer transformer layers to optimize computational load and predictive accuracy.


\begin{figure*}[ht]
    \centering
    \includegraphics[width=0.99\linewidth]{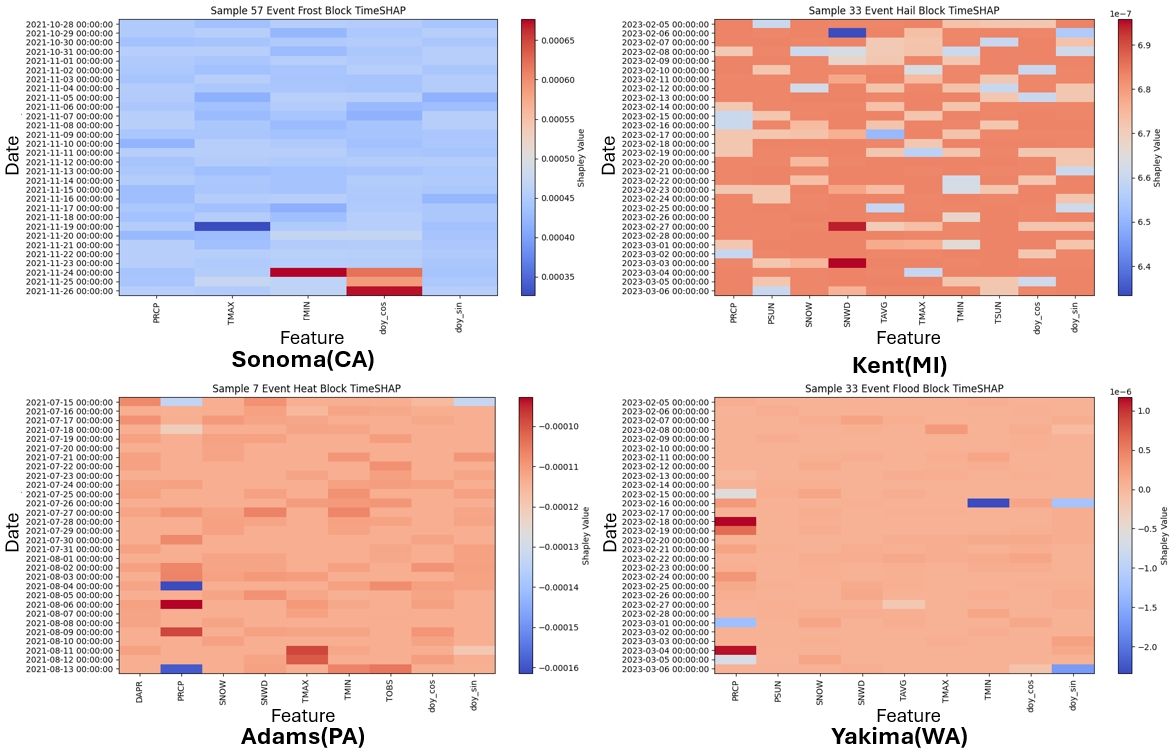}
    \caption{Local explainability heatmaps depicting the temporal and spatial variability in feature significance for distinct extreme weather events across selected agricultural counties. The color intensity indicates the magnitude of feature influence, reflecting the meteorological dynamics specific to each hazard type and location.}
    \label{fig:local}
\end{figure*}

Explainability of our framework is ensured through both local and global explanatory methodologies. At the local scale, TimeSHAP is employed to determine temporal feature significance using Shapley values, identifying how each feature at specific time points contributes to individual predictions. This method provides nuanced, instance-specific insights into model decision-making processes. Recognizing the limitations of directly aggregating local Shapley values due to their variability, we propose a novel global explanatory method. First, local TimeSHAP results generate two-dimensional importance matrices which are indexed temporally and by feature. Second, critical temporal segments are identified using either the magnitude of Shapley values or attention weights derived from the forecasting step. Finally, global feature relevance is calculated by summing these temporal and feature-specific importance measures. This global matrix offers a clearer view of which features tend to matter most, and when, across different hazard types and regions. For example, it can reveal recurring patterns such as rising temperatures in early spring being a consistent precursor to frost risk. Such insights support more informed, proactive responses by highlighting not just what the model has learned, but how its reasoning aligns with agronomic understanding. By integrating both instance-level specificity and population-level regularities, the dual-layered approach makes the model’s decision process more transparent and actionable for stakeholders to evaluate predictions and agricultural risks.

\section{Result}
\label{sec:results}
The forecasting performance of sequential models across the four selected agricultural counties is presented in Table~\ref{tab:predict}. Given regional differences in meteorological data availability, we implemented region-specific models instead of a single universal approach. The results indicate particularly strong forecasting accuracy in Pennsylvania, with the BiLSTM model achieving the lowest errors (MAE: \textbf{0.0201}, RMSE: \textbf{0.0704}). In contrast, higher forecasting errors emerged in Washington, largely due to difficulties accurately predicting Flood events (MAE ~0.30) and Heat events (MAE ~0.10). Nonetheless, the sequential models demonstrated robust predictive capabilities overall, with BiLSTM models frequently outperforming Transformer models. Based on these outcomes, we selected the optimal model for each region to inform subsequent explainability analyses.

\begin{table}[t]
\centering
\captionsetup{width=0.9\columnwidth} 
\caption{Average model performance over all selected extreme events on unseen samples}
\label{tab:predict}
\resizebox{0.95\columnwidth}{!}{
\begin{tabular}{@{}lllllll@{}}
\toprule
\multirow{2}{*}{} & \multicolumn{2}{|l|}{LSTM} & \multicolumn{2}{|l|}{BiLSTM} & \multicolumn{2}{|l|}{Transformer} \\ \cmidrule(l){2-7} 
 & MAE & RMSE & MAE & RMSE & MAE & RMSE \\ \midrule
Sonoma(CA) & 0.0562 & 0.1642 & \textbf{0.0442} & 0.1649 & 0.0608 & \textbf{0.1629} \\
Kent(MI) & 0.0365 & 0.1697 & \textbf{0.0313} & \textbf{0.1692} & 0.0360 & 0.1701 \\
Adams(PA) & 0.0261 & 0.0711 & \textbf{0.0201} & \textbf{0.0704} & 0.0308 & 0.0716 \\
Yakima(WA) & 0.0758 & 0.2901 & 0.0683 & 0.2941 & \textbf{0.0652} & \textbf{0.2791} \\ \bottomrule
\end{tabular}
}
\end{table}

\begin{figure*}[ht]
    \centering
    \includegraphics[width=0.99\linewidth]{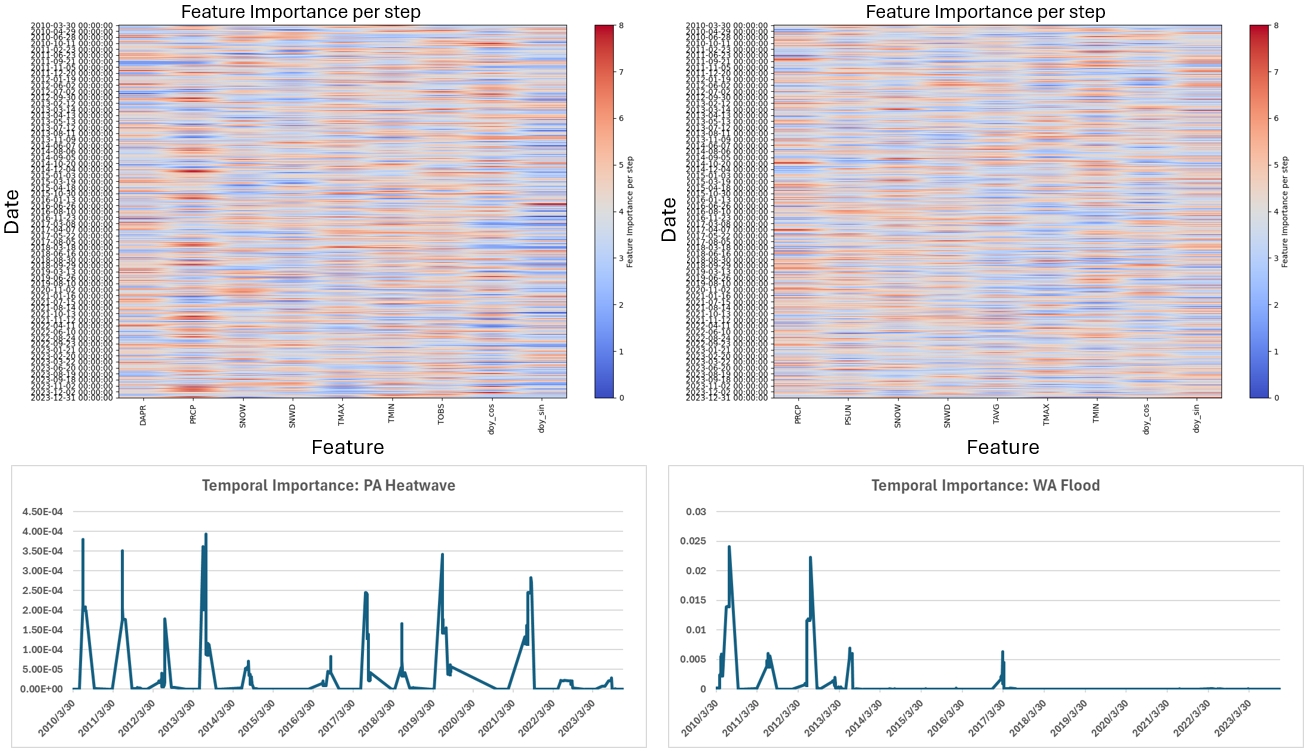}
    \caption{Aggregated global explainability analysis for California and Michigan, highlighting distinct seasonal patterns and the dominant climatic features underpinning frost and hail event predictions. The analysis emphasizes temperature and precipitation dynamics, aligning closely with recognized meteorological processes.}
    \label{fig:global}
\end{figure*}
Figure \ref{fig:local} visualizes local explainability analyses for individual extreme weather events, clearly illustrating both temporal and spatial variations. The differing magnitudes within the heatmaps reflect shifts in the temporal dynamics and significance of meteorological features throughout event periods. Furthermore, distinct feature prominence across regions highlights unique local climatic conditions influencing extreme event profiles. For instance, Sonoma County (California) experiences pronounced temporal shifts in minimum (TMIN) and maximum (TMAX) temperatures during frost events, directly corresponding to frost formation mechanisms. In Kent County (Michigan), hail events exhibit variable importance for snow depth (SWND), suggesting its potential influence on near-surface thermal dynamics. Adams County (Pennsylvania) prominently features daytime maximum temperatures and precipitation (PRCP) during heat events, capturing the agricultural impact of prolonged dry spells. In Yakima County (Washington), flood events underscore precipitation and soil moisture levels, consistent with the region’s susceptibility to sustained rainfall and flooding. These distinctions align well with established meteorological principles, offering clear guidance for targeted agricultural risk mitigation.

Figure \ref{fig:global} provides aggregated global explanations for California and Michigan, elucidating how regional climate patterns influence frost and hail events. In California, a pronounced seasonal rhythm emerges, with meteorological feature importance peaking sharply during colder months, coinciding precisely with typical frost conditions. Minimum and maximum temperatures, along with precipitation, emerge consistently as key predictive features. These findings resonate with known meteorological processes, notably nocturnal radiative cooling and saturation conditions critical for frost formation. In Michigan, however, the analysis highlights different seasonal periodicities associated with hail events. Here, precipitation and snow depth emerge as the dominant indicators. Heavy precipitation marks strong convective storm activity, which is vital for hail development. Additionally, snow depth potentially affects surface temperature gradients, influencing boundary-layer conditions conducive to storm formation during seasonal transitions. This alignment of model interpretations with recognized meteorological dynamics underscores their credibility.

Combining local and global explanations, alongside with feature-level and temporal perspectives, provides a comprehensive framework for interpreting extreme event forecasts. This integrated approach is crucial for unified multi-hazard forecasting, as it captures the intricate relationships of meteorological features across varying temporal scales and spatial contexts. By addressing both immediate and longer-term climatic influences, enhanced explainability significantly aids stakeholder decision-making, enabling precise interventions, efficient resource management, and improved disaster risk mitigation strategies tailored to diverse agricultural environments.

\section{Conclusion}
\label{sec:con}
This study introduces a unified multi-agricultural hazard forecasting system that effectively integrates deep learning models with robust temporal explainability techniques. Our approach demonstrates high predictive accuracy for multiple concurrent climate hazards across diverse agricultural contexts. By combining attention mechanisms and TimeSHAP, the proposed system not only pinpoints key predictive features but also reveals critical timing in the buildup to hazardous events. These insights empower stakeholders, such as farmers and policymakers, to implement timely and targeted mitigation strategies. Ultimately, the integration of explainability within the forecasting model fosters greater transparency and stakeholder confidence, enhancing practical resilience to climate-driven agricultural risks.


\bibliographystyle{elsarticle-harv} 
\bibliography{cas-refs}

\end{document}